\documentclass[journal]{IEEEtran}

\usepackage{cite}
\usepackage{xcolor}
\usepackage{amsmath}
\usepackage{amssymb}
\usepackage{textcomp}
\usepackage{times}
\usepackage{tikz}
\usetikzlibrary{arrows,shapes,shadows,positioning,fadings,decorations,decorations.pathmorphing,matrix}
\tikzstyle{block}=[draw opacity=0.7,line width=1.4cm]
\newcommand{\node}{v}

\newcommand{\labelfun}{L}

\newcommand{\hash}{\pi}

\newcommand{\path}{p}
\usepackage{multirow}
\usepackage[linesnumbered,vlined]{algorithm2e}
\SetVlineSkip{0pt} 

\begin{document}
\title{{A tree-based kernel for graphs with continuous attributes}}

\author{Giovanni Da San Martino,
Nicol\`o Navarin, {\it Member, IEEE}, and
Alessandro Sperduti, {\it Senior Member, IEEE}
\thanks{G. Da San Martino is with the ALT research group, Qatar Computing Research Institute, HBKU, P.O. Box 5825 Doha, Qatar.}
\thanks{N. Navarin and A. Sperduti are with the Department of Mahematics, University of Padova, via trieste 63, Padova, Italy.}
\thanks{\textbf{This work has been submitted to the IEEE for possible publication. Copyright may be transferred without notice, after which this version may no longer be accessible.}}}


\maketitle

\begin{abstract}
The availability of graph data with node attributes that can be either discrete or real-valued is
constantly increasing. 
While existing kernel methods are effective techniques for dealing with graphs having discrete node labels, 
their adaptation to non-discrete or continuous node attributes has been limited, mainly for computational issues. 
Recently, a few kernels especially tailored for this domain, and that trade predictive performance 
for computational efficiency, have been proposed.
In this paper, we propose a graph kernel for complex and continuous nodes' attributes, 
whose features are tree structures extracted from specific graph visits. 
The kernel manages to keep the same complexity of state-of-the-art kernels while implicitly using a larger feature space. 
We further present an approximated variant of the kernel which reduces its complexity significantly. 
{Experimental results obtained on six real-world datasets show that the kernel is the best performing one on most of them.
Moreover, in most cases the approximated version reaches comparable performances to current state-of-the-art kernels in terms of classification accuracy 
{while greatly shortening the running times}. }

%
\end{abstract}

\section{Introduction}
There is an increasing availability of data in the form of attributed graphs, 
i.e. graphs where some information is attached to nodes and edges (and to the graph itself).
For computational reasons, available machine learning techniques for graph-structured data have been focusing on problems whose data can be modelled as graphs with discrete attributes. 
However, in many application domains, such as bioinformatics and action recognition, non-discrete node attributes are available~\cite{Borgwardt2005a, Karaman2014}. 
For example, many bioinformatics problems deal with proteins.
It is possible to represent a protein as a graph, where nodes represent secondary structure elements. Two nodes are connected whenever they are neighbors either in the amino acid sequence or in space~\cite{Borgwardt2005a}.
Each node {has} a discrete-valued attribute indicating the structure it belongs to (helix, sheet or turn). 
Moreover, several chemical and physical measurements can be associated with each node, such as the length of the secondary structure element in \AA, its hydrophobicity, polarity, polarizability etc. 
In some tasks, discarding such type of information has a significantly negative impact on the predictive performance (see section~\ref{sec:experiments}). 
%

Most of the graph kernels in the literature are not suited for non-discrete node labels since their computational efficiency hinges on hard matches between discrete labels. 
Of course, this strategy cannot work for  non-discrete labels, which in general
are all distinct. An alternative would be to define a kernel function between graph nodes,  however the resulting computational times become unfeasible, as in the case \mbox{of \cite{Kriegel05shortestpath}}. 
For this reason, recently there has been an increasing interest in the definition of graph kernels that can efficiently deal with continuous-valued attributed graphs. The problem is challenging because both fast and expressive kernels (in terms of discriminative power
) are looked for.

In this paper, we present a new kernel inspired by the graph kernel framework proposed in \cite{Dasan2012}. 
The features induced by the kernel are tree structures extracted from breadth-first visits of a graph (contrary to \cite{Borgwardt2005a} an edge is only traversed once per visit). 
 We extend the definition of tree kernels, and consequently derive a graph kernel, which is able to deal with complex and continuous node labels. 
 As the experimental results show, the richer feature space allows to reach state-of-the-art classification performances on real world datasets. 
 While the computational complexity of our kernel is the same as competing ones in literature, 
 we describe an approximated computation of the kernel between graph nodes in order to reach 
 %
 {lower} running times, while keeping state-of-the-art results. 
%
\section{Notation}\label{sec:notation}
A graph \mbox{$G=(V_G,E_G,\labelfun_G)$} is a triplet where $V_G$ is the set of $n$ vertices, $E_G$ the set of $m$ edges and $\labelfun_G()$ a function mapping nodes to discrete labels. 
A graph is undirected if $(\node_i,\node_j)\in E_G \Leftrightarrow (\node_j,\node_i)\in E_G$, otherwise it is directed. 
{A path $\path(\node_i, \node_j)$ of length $s$ in a graph $G$ is a sequence of nodes $u_0,\ldots,u_{s-1}$, where $u_i \in V_G$, $u_0=\node_i$, $u_s=\node_j$ and $(u_k,u_{k+1})\in E_G$ for $0\leq k< s-1$. }
A cycle is a path for which {$u_0=u_{s-1}$.}
A graph is acyclic if it has no cycles.
A tree is a directed acyclic graph where each node has exactly one incoming edge, except the root node which has no incoming edge. 
The root of a tree $T$ is represented by $r(T)$. 
The i-th child (outgoing edge) of a node $v \in V_T$ is referred to as $ch_v[i]$.
The number of children of a node $v$ is referred to as $\rho(v)$ ($\rho$ is the maximum out-degree of a tree or graph). 
A leaf is a node with no children. 
A proper subtree rooted at node $\node$ comprises $\node$ and all its descendants. 
\section{Related work}\label{sec:related}
In the last few years, several graph kernels for discrete-valued graphs have been proposed in literature. 
Early works presented kernels that have to be computed in closed form, such as the \textit{random walk} kernel~\cite{Gartner2003a} 
or the \textit{shortest path} kernel~\cite{Kriegel05shortestpath}.
These kernels suffer from a relatively high computational complexity: $O(n^3)$ and $O(n^4)$ respectively.
More recently, research focused on the efficiency of kernel calculation.
State-of-the-art kernels use explicit feature mapping techniques~\cite{Costa2010,NIPS2009_0533,Dasan2012}, with computational complexities almost linear in the size of the graphs.
If we consider {graphs with continuous-valued labels}, this last class of kernels cannot be easily modified to deal with them {because
their efficiency hinges on the ability to perform computation only for discrete labels that match. Of course, this is not possibile when considering continuos-valued labels}. Between the two {obvious} possible solutions, i.e. adopt slower kernels or discretizing/ignoring the continuous attributes of the graphs, the latter approach was usually the preferred one~\cite{Birlinghoven}.
In \cite{Kriege2012} a kernel for graphs with continuous-valued labels has been presented. The kernel matches common subgraphs up to a fixed size $k$, and has complexity $O(n^k)$.

In \cite{Feragen2013} another more efficient kernel has been presented.
This kernel is a sum of path kernels, that in turn are a sum of node kernels.
The computational complexity of the kernel is $O(n^2(m+\log n + d +\sigma^2))$ where $n$ and $m$ are the number of nodes and edges in the graph, respectively, $\sigma$ is the depth of the graph and $d$ is the dimension of the vectors associated to nodes.
However, experimental results show that this kernel cannot achieve the same predictive performance as other computationally more demanding graph kernels, e.g. the  \textit{shortest path} kernel. 

Very recently, two kernel frameworks able to deal with continuous and vectorial labels have been proposed: in \cite{Neumann2015} authors propose to use Locality Sensitive Hashing to discretize continuous and vectorial labels, while in \cite{Orsini} a very general framework of graph kernels is proposed.

The experience on discrete-labeled graphs teaches us that path-features are not the most expressive ones. 
In fact, in \cite{NIPS2009_0533,Dasan2012} it is shown that tree-features can express a more suitable similarity measure for many tasks. 
The framework presented in~\cite{Dasan2012} is especially interesting since it allows to easily define a kernel for graphs from a vast class of tree kernels, and it constitutes the starting point of our proposal. 

\section{Ordered Decomposition DAG Kernels for Graphs with Discrete Labels} \label{sec:oddkernel}

The kernel we are going to propose is based on tree structures. This section briefly recalls the procedure for extracting them from a graph~\cite{Dasan2012}~\cite{DBLP:journals/ijon/MartinoNS16}. 
 In order to map the graphs into trees, two intermediate steps are needed:

\begin{figure}
\centering
\begin{tikzpicture}[auto,thick,scale=.45]
   \tikzstyle{graph}=[scale=1.0,minimum size=12pt, inner sep=0pt, outer sep=0pt,fill=blue,circle, text=white] 
		\node (s) [graph] {s}   
					child {
            node (b) [graph] {b}
        	};
         \node (e) [graph] at (0:2){e}   
         		child { node (d) [graph] {d}};		
    ;    	
    \draw
    (s) to (e)
    (b) to (d)
    (b) to (e)
    ;
    
  \node [below, right, xshift=-0.5cm,yshift=-0.75cm] at (b) {Input Graph};  
  \small
	 \draw [red,->,line width=5pt] (4,-0.8)--(6,-0.8);
	 \node[circle,scale = 0.65,color=red,draw,below, right, xshift=1.25cm,yshift=0.15cm] at (e) {\bf  1};
   \tikzstyle{node}=[graph]
        \node [node,xshift=3.3cm, yshift=0cm] {\textbf{s}} [->] 
                child {
                    node [node] {\textbf{e}}[] 
                    child {node (d2)[node, xshift=0.35cm]{\textbf{d}}[]}
                }
                child {
                    node (b2) [node]{\textbf{b}}[]
                }
				;
         \draw [->]
         (b2) to (d2)
        ;
        \node (e) [node,xshift=6.5cm,yshift=0cm] {e}
        		child[->] {
            			node (s) [node] {s}
            			}
						child[->]{
									node (b) [node] {b}
									}
	   				child[->] {
            			node (d) [node] {d}
            			}
        ;
        \node [node,xshift=6.5cm,yshift=-1.2cm] {b}
        		child[->] {
            			node [node] {s}
            			}
						child[->]{
									node [node] {e}
									}
	   				child[->] {
            			node [node] {d}
            			}
        ;    
        \node [node,xshift=4.5cm, yshift=0cm] {\textbf{d}} [->] 
                child {
                    node [node] {\textbf{b}}[] 
                    child {node (d2)[node, xshift=0.35cm]{\textbf{s}}[]}
                }
                child {
                    node (e2) [node]{\textbf{e}}[]
                }
				;
         \draw [->]
         (e2) to (d2)
        ;     
         
 \node [below, left, xshift=0.75cm,yshift=-1.5cm] at (e2) {DAG visits};  
   
  \draw [red,->,line width=5pt] (15.5,-5)--(15.5,-7);  
	 \node[circle,scale = 0.65,color=red,draw,below, right, xshift=2.3cm,yshift=-1.9cm] at (d2) {\bf  2};
   
           \node [node,xshift=4.5cm, yshift=-3.9cm] {\textbf{s}} [->]
                child {
                    node (x2)[node] {\textbf{b}}[] 
                    child {node (d2)[node, xshift=0.35cm]{\textbf{d}}[]}
                }
                child {
                    node (b2) [node]{\textbf{e}}[]
                }
				;
         \draw [->]
         (b2) to (d2)
        ;
        \node [above, left, xshift=0.2cm,yshift=0.43cm] at (x2) {\scriptsize 1};  
        \node [above, left, xshift=0.2cm,yshift=0.43cm] at (b2) {\scriptsize 2};  
        \node [above, left, xshift=-0.15cm,yshift=0.35cm] at (d2) {\scriptsize 1};  
        \node [above, left, xshift=0.6cm,yshift=0.35cm] at (d2) {\scriptsize 1};  

        \node (e) [node,xshift=6cm,yshift=-3.3cm] {e}
        		child[->] {
            			node (s) [node] {b}
            			}
						child[->]{
									node (b) [node] {d}
									}
	   				child[->] {
            			node (d) [node] {s}
            			}
        ;
        
        \node [above, left, xshift=0.35cm,yshift=0.40cm] at (s) {\scriptsize 1};  
        \node [above, left, xshift=0.05cm,yshift=0.35cm] at (b) {\scriptsize 2};  
        \node [above, right, xshift=-0.35cm,yshift=0.40cm] at (d) {\scriptsize 3};

        \node [node,xshift=6.1cm,yshift=-4.5cm] {b}
        		child[->] {
            			node (s)[node] {d}
            			}
						child[->]{
									node (b)[node] {e}
									}
	   				child[->] {
            			node (d)[node] {s}
            			}
        ;   
        \node [above, left, xshift=0.35cm,yshift=0.40cm] at (s) {\scriptsize 1};  
        \node [above, left, xshift=0.05cm,yshift=0.35cm] at (b) {\scriptsize 2};  
        \node [above, right, xshift=-0.35cm,yshift=0.40cm] at (d) {\scriptsize 3};  
 
        \node [node,xshift=7.5cm, yshift=-3.7cm] {\textbf{d}} [->] 
                child {
                    node (x2)[node] {\textbf{b}}[] 
                    child {node (d2)[node, xshift=0.35cm]{\textbf{s}}[]}
                }
                child {
                    node (e2) [node]{\textbf{e}}[]
                }
				;
         \draw [->]
         (e2) to (d2)
        ;    
        \node [above, left, xshift=0.2cm,yshift=0.43cm] at (x2) {\scriptsize 1};  
        \node [above, left, xshift=0.2cm,yshift=0.43cm] at (e2) {\scriptsize 2};  
        \node [above, left, xshift=-0.15cm,yshift=0.35cm] at (d2) {\scriptsize 1};  
        \node [above, left, xshift=0.6cm,yshift=0.35cm] at (d2) {\scriptsize 1};  
  
   \node (odlabel) [below, left, xshift=-1.45cm,yshift=-1.5cm] at (e2) {Ordered DAGs};  
  
   \draw [red,->,line width=5pt] (7.5,-11)--(5.5,-11);
 	 \node[circle,scale = 0.8,color=red,draw,below, right, xshift=-0.3cm,yshift=0.5cm] at (7,-11) {\bf  3};

 \node [above, right, xshift=-4cm] at (odlabel) {FeatureMap};  
  
   \draw [->,line width=1pt] (10.8,-7)--(11.8,-8);
   \node [xshift=5cm,yshift=-3.0cm]  {\scriptsize child order};  
     	    \node [node, xshift=0.9cm,yshift=-3.7cm] {\textbf{s}} [->] 
                child {
                    node [node] (b3){\textbf{b}}[] 
                    child {node (d3)[node, xshift=0.0cm]{\textbf{d}}[]}
                }
                child {
                    node (e3) [node]{\textbf{e}}[]
                    child {node (d4)[node, xshift=0.0cm]{\textbf{d}}[]}
                };
	  ;
	          \node [above, left, xshift=0.2cm,yshift=0.43cm] at (b3) {\scriptsize 1};  
	         \node [above, left, xshift=0.2cm,yshift=0.43cm] at (e3) {\scriptsize 2};  
	         \node [above, left, xshift=0.0cm,yshift=0.43cm] at (d3) {\scriptsize 1}; 
	         \node [above, left, xshift=0.0cm,yshift=0.43cm] at (d4) {\scriptsize 1};   
           \node [node, xshift=0cm,yshift=-3.7cm] {\textbf{e}} [->]
           child {node (d5)[node, xshift=0.0cm]{\textbf{d}}[]};
           	\node [above, left, xshift=0.0cm,yshift=0.43cm] at (d5) {\scriptsize 1};   
           \node [node, xshift=1.85cm,yshift=-3.7cm] {\textbf{b}} [->]
           child {node (d6)[node, xshift=0.0cm]{\textbf{d}}[]};
          \node [above, left, xshift=0.0cm,yshift=0.43cm] at (d6) {\scriptsize 1};   
           \node [node, xshift=2.75cm,yshift=-3.7cm] {\textbf{d}};

\matrix (m)  [matrix of nodes,nodes in empty cells,yshift=-3.0cm,xshift=1.6cm,text height=1em,text width=.8em]
{
 1 & ... & 1 & ... & 1 & ... & 4 & ...  \\
  };
  \foreach \j in {1,...,8}{
  \draw (m-1-\j.north west) rectangle (m-1-\j.south east);
  }
\end{tikzpicture}
\caption{ODD kernel summary. 1) Decomposition of a graph into its DDs; 2) definition of a total ordering among the children of each node; 3) generation of an explicit FeatureMap extracting all proper subtrees (ST kernel) from the set of Ordered DDs.\label{fig:oddsummary}}
\end{figure}
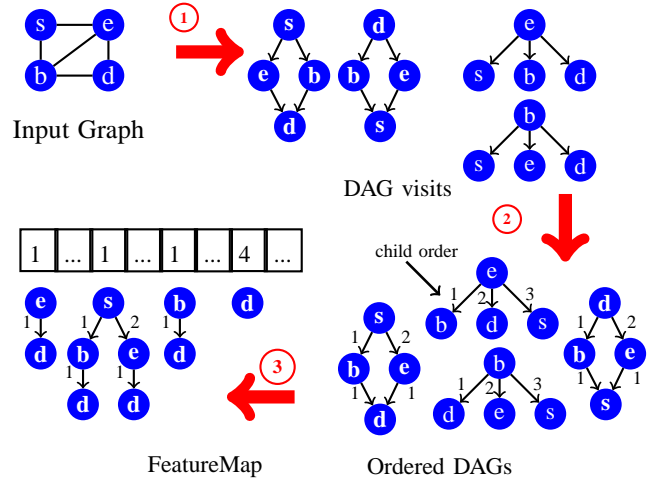

\begin{enumerate}
 \item map the graph $G$ into a multiset of Decomposition DAGs $DD_G=\{DD_{G}^{\node_{i}} | \node_{i} \in V_{G} \}$, where $DD_{G}^{\node_{i}}$ is formed by the nodes and the directed edges in the shortest path(s) between $\node_i$ and any $\node_j\in V_G$. 
 Figure~\ref{fig:oddsummary}-1) shows an example of $DD_G$. 
 In order to reduce the time required for evaluating the kernel, the visits can be restricted to those nodes whose shortest path length w.r.t. $\node_i$ is not greater than a parameter $h$. 
 {We recall the following facts discussed in more details in~\cite{Dasan2012}. }
 {Let $H_{max}$ be the maximum number of nodes of a $DD^{\node_i}_{G}\in DD_G$, then $H_{max}\leq \lfloor \frac{\rho^{h+1}+1}{\rho-1}\rfloor \leq n$. } 
 The decomposition we have defined ensures that isomorphic graphs are represented by the same multiset of DAGs, 
 {which is a necessary condition for the kernels we will propose to be well-defined. } 
 The computation of the multiset $DD_G$ for a graph G requires $O(nm)$ time. 
  \item Since the kernel we are going to describe in this paper requires the DAG nodes to be ordered, 
   a strict partial order between nodes in $DD_{G}^{\node_{i}}$ has been defined yielding an Ordered Decomposition DAG $ODD_{G}^{\node_{i}}$. 
{The ordering relies on an encoding of the proper sub-DAGs rooted at each node as strings. 
Let $\kappa: \sum^*\rightarrow\sum^w$ and $\hash:V\rightarrow\sum^w$ be two functions returning strings of length $w$. 
We assume $\kappa$ to be a perfect hash function for a sufficiently large $w$. 
Let $||$ be the concatenation operator between strings. 
We then define the encoding function for a node $\hash(\cdot)$ as: 
\begin{equation}
 \hash(\node)=\kappa\big(\kappa(L(\node))||\hash(ch_\node[1])||\ldots||\hash(ch_\node[\rho(\node)])\big). 
 \label{eq:nodeencoding}
\end{equation}
When a node $\node$ is a leaf, eq.~\eqref{eq:nodeencoding} reduces to $\hash(\node)=\kappa(L(\node))$. 
The fact that the output of $\hash$ is of fixed size, and that it is a combination of values returned by a perfect hash function, ensures that it is well-defined. 
A strict partial order between nodes is then the alphanumeric ordering between strings $\hash(\cdot)$. }

{Although there exist different DAGs represented by the same $\hash(\cdot)$ values, the swapping of nodes with the same $\hash(\cdot)$ value does not change the feature space representation of the examples~\cite{Dasan2012}. 
}
    Figure~\ref{fig:oddsummary}-2) shows an example of such ordering. 
    Once $\kappa(\cdot)$ values are computed, ordering the sibling of each node in a DD requires $\sum_{\node\in DD}\rho(\node)\log\rho(\node)\leq\log\rho\sum_{\node\in DD}\rho(\node)=m \log \rho$ steps. Ordering all $n$ DAGs in a $DD_G$ requires $O(nm \log \rho)$ time. 
\item Finally, any Ordered DAG (ODD) is mapped into a multiset of trees. 
 Let us define $T(\node_i)$ as the tree resulting from the visit of $ODD_{G}^{\node_{i}}$ starting from node $\node_i$: the visit returns 
 the nodes reachable from $\node_{i}$ in $ODD_{G}^{\node_{i}}$. 
 If a node $\node_{j}$ can be reached more than once, more occurrences of $\node_{j}$ will appear in $T(\node_{i})$.
 In the following, the notation $T_l(\node_i)$ indicates a tree visit of depth $l$.
 
  Notice that any node $\node_j$ of the DAG having $l>1$ incoming edges will be duplicated $l$ times in $T(\node_i)$ (together with all the nodes that are reached from $\node_j$ in the same visit). 
Thus, given a DAG $ODD^{\node_{i}}_{G_1}$, the total number of nodes of all the tree visits, i.e. $\sum_{\node\in ODD^{\node_{i}}_{G_1}}|T(\node)|$, can be exponential with respect to $|V_{G_1}|$. However, such observation does not imply that the complexity of the resulting kernel is exponential since tree visits need not to be explicitly computed to evaluate the kernel function. 
\end{enumerate}
The Ordered Decomposition DAGs (ODD) kernel we are going to use in this paper, is defined as: 
  \begin{equation}
   K(G_1,G_2) = \!\!\!\!\!\!\!\!\!\!\!\!\displaystyle \sum_{\substack{OD_1 \in ODD_{G_1} \\OD_2 \in ODD_{G_2}}} \sum_{\substack{\node_1\in V_{OD_1}\\\node_2\in V_{OD_2}}} \sum_{j=1}^h C(r(T_j(\node_1)),r(T_j(\node_2)))
    \label{eq:k}
  \end{equation}
where $C()$ is a recursive function {used} for computing the Subtree Kernel (ST) \cite{Moschitti2006, DBLP:conf/nips/ViswanathanS02}:   
{\begin{equation}
C_{ST}(\node_1, \node_2) = 
\begin{cases}
\lambda  \hspace{0.9cm}\text{if $L(\node_1)=L(\node_2) \wedge \node_1, \node_2$ are leaves} \\
\lambda\displaystyle \prod_{j=1}^{\rho(\node_1)} C_{ST}(ch_{\node_1}[j] , ch_{\node_2}[j]) \vspace{-0.3cm}\\
\hspace{1.5cm}\text{if $L(\node_1)=L(\node_2)\wedge\rho(\node_1)=\rho(\node_2)$} \vspace{0.3cm}\\
 0  \hspace{4.7cm}\text{otherwise.}
\end{cases}
 \label{eq:cste}
\end{equation}
Here $\lambda$ is a kernel parameter. }
The ST kernel counts the number of {matching} proper subtrees between the two input trees. 
While we focus on the ST kernel in this paper, similar extensions can be easily applied to other tree kernels. 

\section{Graph Kernels for Continuous Node Labels} \label{sec:stcontinuous}
This section extends the ST kernel to deal with non-discrete node labels. 
However, the kernel we describe is also able to deal with continuous labels only. 
Let us now extend a few definitions to the continuous domain. 
In order to simplify the presentation, we will also cast the notation and the following function definitions to the domain of the tasks we address in the experimental section. 
Let us define a graph with continuous attributes as $G=(V_G,E_G,L_G,A_G)$ where $A_G()$ is a function associating to each node a real-valued vector in $\mathbb{R}^d$. 
In the following, we will assume the DAGs to be ordered as described in Section~\ref{sec:oddkernel}. 
Let us assume a kernel on continuous attributes $K_A(\node_1,\node_2)$ is given as parameter.
We start in section~\ref{sec:extensions} by describing a straightforward extension of the kernel for discrete labels presented in Section~\ref{sec:oddkernel}. 
We then propose an alternative kernel definition in section~\ref{sec:ourproposal}. Moreover, in the same section, we provide an efficient algorithm for computating the kernel. 

\subsection{A First Kernel For Graphs With Continuous Node Labels} \label{sec:extensions}
A straightforward way to extend the ST kernel to deal with continuous labels is to introduce the $K_A()$ kernel on continuous labels {wherever the two discrete labels match: 
\begin{equation}
C'_{ST}(\node_1, \node_2) = 
\begin{cases}
\lambda \cdot K_A(v_1,v_2) \hspace{1.3cm}\text{if $L(\node_1)=L(\node_2) \wedge$}\\ \hspace{3.6cm}\text{$\node_1,\node_2$ are leaves}\\ 
\vspace{-0.3cm}\lambda \cdot K_A(v_1,v_2) \displaystyle \cdot\prod_{j=1}^{\rho(\node_1)}\hspace{-0.15cm}C'_{ST}(ch_{\node_1}[j] , ch_{\node_2}[j])\vspace{0.2cm} \\ \hspace{1.3cm}\text{if $L(\node_1)=L(\node_2) \wedge$ $\rho(\node_1)=\rho(\node_2)$}\vspace{0.2cm}\\
 0  \hspace{4.4cm}\text{otherwise.}
\end{cases}
 \label{eq:cstprime}
\end{equation}
}
A drawback of this approach is that the kernel value between $v_1$ and $v_2$ may be influenced by the 
function used for ordering the nodes.
For example, assume that $v_1$ and $v_2$ have the same number of children, each one with the same discrete label, but with a different continuous label. 
In this case the pairs for which $K_A$ is computed, and consequently the value of the kernel evaluation, depends on how identical discrete labels are ordered. 
Even if we extend the ordering function to consider continuous labels, the selection of pairs would be biased by the ordering function. 
Ideally we would like to compute the kernel for all those nodes whose discrete labels are identical. 
However, extending eq.~\eqref{eq:cstprime} with such goal would dramatically increase its complexity. 
 
\subsection{Our Proposal} \label{sec:ourproposal}

We define now an efficient extension of the ST kernel which, given a well defined ordering as the one described in setion~\ref{sec:oddkernel}, is not sensitive to the disposition of the nodes with identical discrete labels. 
The way we propose to extend the $C()$ function of the ST kernel 
to handle complex node labels is the following: 
{\begin{equation}
C_{CST}(\node_1,\node_2)=
\begin{cases}
\lambda \cdot K_{A}(\node_1,\node_2) \hspace{1.3cm}\text{if $L(\node_1)=L(\node_2)\wedge$} \\
 \hspace{3.2cm}\text{$\node_1, \node_2$ are leaf nodes} \\
{\lambda\cdot} K_{A}(\node_1,\node_2) \cdot C_{ST}(\node_1, \node_2) \\ \hspace{1.2cm}\text{if $L(\node_1)=L(\node_2)\wedge$ $\rho(\node_1)=\rho(\node_2)$} \\ 
0  \hspace{4.6cm}\text{otherwise}\\
\end{cases}\label{eq:cst}
\end{equation}}

Note that the rightmost quantity on the right-hand side of case 2 in the equation is the original $C_{ST}$ function in eq.~\eqref{eq:cste}, i.e. it does not consider continuous labels.
Thus, the kernel for complex node labels is applied only to the
root node of the tree-features. However, if $C_{ST} (v_1 , v_2 ) > 0$
then $C_{ST} (v_1', v_2' ) > 0$ for all the pairs of nodes $(v_1' , v_2')$ for
which $C_{ST}()$ is recursively evaluated during the computation
of $C_{ST}(v_1 , v_2)$. As a consequence, recalling that eq. (5) is
used in combination with eq. (2), when $C_{CST}(v_1 , v_2)$ is
evaluated, the kernel $K_A()$ is surely evaluated as well. Thus, the continuous 
labels belonging to all the nodes of a matching tree-feature contribute to the determination of the kernel value. 
Since $K_A$ is only evaluated on root nodes and eq.~\eqref{eq:cst} is evaluated on all pairs of nodes of the input graphs, the outcome of the kernel is clearly independent of the disposition of (sibling) nodes with identical discrete labels. 
%
%
%
{Eq.~\eqref{eq:cst} defines a positive semidefinite kernel since it is the product of positive semidefinite kernels. 
By using the well-defined ordering in~\eqref{eq:nodeencoding}, eq.~\eqref{eq:k}, instantiated with eq.~\eqref{eq:cst}, is a valid kernel as showed in~\cite{Dasan2012}. }
We call it ODDCL$_{ST}$ kernel. 
{In the following we instantiate the kernel on vectorial attributes as the gaussian kernel: $K_A(v,v')=e^{-\beta ||A(v)-A(v')||^2}$ (here $\beta$ is a kernel parameter) since that is kernel we will use in the experimental section. }

Eq.~\eqref{eq:cst} has the only purpose to show how the computation of the ST kernel changes from the discrete to the complex node label domain. 
The algorithm we are going to use to compute the kernel is more efficient than the direct evaluation of eq.~\eqref{eq:cst} 
and it is based on a fast algorithm for computing the ST kernel for discrete node labels~\cite{DBLP:conf/nips/ViswanathanS02}. 
%
The kernel has been implemented in Python and the algorithms presented in this section are (simplified) snippets of the actual code.
\begin{algorithm}
\caption{Sketch of an algorithm to compute the FeatureMap of a graph. The notation is Python-style: \{\} is an HashMap, and the \textbf{in} operator applied to an HashMap performs the lookup of the element in it.\label{alg:FeatureMap}}
\small
\SetKwFunction{algo}{computeFeatureMap}
  \SetKwProg{myalg}{def}{}{}
  \myalg{\algo{G,h}}{
\KwData{$G$= a graph}
\KwData{$h$= maximum depth of the considered structures}
\KwResult{FeatureMap=\{subtreeID:\{veclabels:freq\}\}}
\KwResult{SizeMap=\{subtreeID:size\}}
DDs=computeDecompositionDAGs($G$,$h$)\;
ODDs=order(DDs)\;
FeatureMap=\{\}\;
\For{\rm{ODD} \textbf{in} \rm{ODDs}}{    
  \For{\rm{v} \textbf{in} topologicalSort(ODD)}{
    \For{\rm{j} \textbf{in} 0\dots $h$}{
        subtreeID=encode(T$_j$(v))\;
        SizeMap[subtreeID]=$|$T$_j$(v)$|$\;
        \eIf{ \rm{subtreeID} \textbf{not in} \rm{FeatureMap}}{

           FeatureMap[subtreeID]=\{v:1\}\;
        }{ 
            \eIf{\rm{v} \textbf{not in} FeatureMap[subtreeID]}{
                    FeatureMap[subtreeID][v]=1\;
	      }{
                FeatureMap[subtreeID][v]+=1\;
              }
          }
  }
}
}
\Return{\rm{FeatureMap, SizeMap}}
}
\end{algorithm}
Algorithm~\ref{alg:FeatureMap} computes the \textit{FeatureMap} of a graph $G$. 
\textit{FeatureMap} is an HashMap that indicizes all the proper subtrees that appear in the graph considering only the discrete node labels. Each subtree encoded by a value $x$ in the FeatureMap, has associated another HashMap containing all the continuos attribute vectors of each subtree encoded by the same value $x$ in the FeatureMap. 
Note that, in our implementation, FeatureMap is a Python dictionary indicized by strings that uniquely encode trees (function \textit{encode} in line 8 of Algorithm~\ref{alg:FeatureMap}). 
We recall that the subtree features do not consider the continuous labels, thus the same subtree may appear multiple times in the same graph.
Each attribute vector has associated its frequency. 
Only subtrees with identical structures and discrete labels are encoded by the same hash value. 
The value of $C_{ST}()$ for two subtrees encoded by the same hash value is $C_{ST}(\node_1, \node_2)=\lambda^{|T(\node_1)|}$ \cite{DBLP:conf/nips/ViswanathanS02}, thus we only need to know the size of the subtree. 
Let us now analyze the computational complexity of the algorithm.
Lines 2-3 require $O(mn)$ and $O(nm \log \rho)$ time, respectively (see Section 3). 
Since the nodes are sorted in inverse topological order, the encoding of line 8 can be computed with a time complexity of $O(\rho)$~\cite{Dasan2012}. 
Let $H \leq \lfloor \frac{\rho^{h+1}+1}{\rho-1}\rfloor \leq n$ be the average number of nodes in a DD (equivalently, $nH$ is the notal number of nodes in all the DDs). 
Lines 8-16 insert a single feature in the $FeatureMap$. The total number of features generated from a graph $G$ 
is then $nH(h+1) \leq (h+1)n^2$ (lines 5-7). 
 The (amortized) cost of inserting all such features in the FeatureMap is $O(nHh)$. 
The overall computational complexity of Algorithm~\ref{alg:FeatureMap} is then $O(n(m \log \rho + hH))$. \
Note that Algorithm~\ref{alg:FeatureMap} has to be executed only once per example. 
\begin{algorithm}
\caption{Sketch of an algorithm for computing the ODDCL$_{ST}$ kernel. $\lambda$ is a kernel parameter.\label{alg:ODDCLkernel}}
\small
\SetKwFunction{algo}{ODDCLkernel}
  \SetKwProg{myalg}{def}{}{}
  \myalg{\algo{$G_1$,$G_2$,$\lambda$,$h$}}{
\KwData{$G_1$,$G_2$=two graphs; $\lambda$= a weighting parameter}
\KwResult{k=the kernel value}
FM1,SM1=computeFeatureMap($G_1$,$h$)\;
FM2,SM2=computeFeatureMap($G_2$,$h$)\;
k=0\;       
\For{\rm{subtreeID} \textbf{in} \rm{FM1}}{ 
            \If{\rm{subtreeID} \textbf{in} \rm{FM2}}{
            \tcc{subtreeID is a feature generated from vertices $v_i \in G_1$ and $v_j \in G_2$ where $C_{ST}(v_i,v_j)\neq 0$}
                \For{\rm{v}$_i$ in FM1[subtreeID]}{\label{line:startapprox}
                   \For{\rm{v}$_j$ in FM2[subtreeID]}{
                        freq1=FM1[subtreeID][v$_i$]\;
                        freq2=FM2[subtreeID][v$_j$]\;   
                        size=SM1[subtreeID]\;                  
                        k+=freq1$\cdot$freq2$\cdot$$\lambda^{\rm{size}}\cdot$K$_A$(v$_i$,v$_j$)
                        \tcc{$C_{ST}(v_i,v_j)=\lambda^{\rm{size}}$} \label{line:endapprox}
                        }
                        }
             }
}
\Return k
}
\end{algorithm}

Algorithm~\ref{alg:ODDCLkernel} sketches the code to compute the kernel value between two graphs.
Once the FeatureMaps have been computed with Algorithm~\ref{alg:FeatureMap}, to calculate the kernel we need to search for matching subtree features in the two FeatureMaps. 
{For any matching subtree feature \textit{subtreeID}, we need to compute the kernel $K_{A}(v_i,v_j), v_i \in V_{G_1}, v_j \in V_{G_2}$ for each pair of vertices that generate the feature \textit{subtreeID} in the two graphs.} 
The complexity of Algorithm~\ref{alg:ODDCLkernel} is linear in the number of discrete features (lines 5-6) and quadratic in the lists of vectorial labels associated to each discrete feature.
Note that there are at most $n$ different attribute vectors in the original graph (one associated to each node), so each discrete feature can be associated with at most $n$ different vectorial labels. 
Thus, the distribution of the $O(hHn)$ elements in a FeatureMap (FM1 or FM2) that maximizes the computational complexity is the one 
having $O(hH)$ different discrete features, each one with an associated list of vectorial labels of size $O(n)$. 
The complexity of Algorithm~\ref{alg:ODDCLkernel} is then $O(hHn^2 Q(K_{A}))$. 

Note that computing eq.~\eqref{eq:cstprime} with Algorithm~\ref{alg:ODDCLkernel} would not be feasible: the kernel $K_A(\node_i, \node_j)$ computed on line 12, which is now computed on a single pair of nodes, should be computed on the whole set of nodes composing the subtrees rooted at $\node_i, \node_j$. 
\subsection{Computation speedup with RBF kernel approximation\label{sec:approximation}}
Profiling the execution of Algorithm~\ref{alg:ODDCLkernel}, the most expensive step is the computation of $K_A()$ for all the pairs of vectorial labels associated to a subtree feature (lines \ref{line:startapprox}-\ref{line:endapprox}). 
In order to speed-up the kernel computation, we propose to approximate this step. 
Recently, \cite{NIPS2008_3495} proposed a method to generate an (approximated) explicit feature space representation for the RBF kernel by Monte Carlo approximation of its Fourier transform. 
This procedure depends on a parameter $D$ determining the size of the approximated feature vector. 
In the following we refer to the approximated feature vector of a node $\node$ as $\hat \phi_{RBF}^D(\node)$; the kernel induced by $\hat \phi_{RBF}^D()$ is positive semidefinite~\cite{NIPS2008_3495}. 
Note that other approximations, such as Nystr\"om~\cite{Williams2001}, can be used. 
Assuming we have one set of identical (w.r.t. discrete labels only) proper subtrees $V_1$ related to a graph $G_1$, and a second set $V_2$ related to a graph $G_2$, we can approximate the computation of the kernel $K_A()$ between the all pairs $(v_i\in V_1, v_j\in V_2)$ as 
\begin{equation}
\sum_{v_i\in V_1} \sum_{v_j\in V_2} k_{RBF}(v_i,v_j) \simeq \langle \sum_{v_i} \hat \phi_{RBF}^D(v_i),\sum_{v_j} \hat \phi_{RBF}^D(v_j)  \rangle.
\label{eq:approx}
\end{equation}
We can now substitute lines 11-17 of Algorithm~\ref{alg:FeatureMap} in order to associate to subtreeID just the sum of the explicit RBF vectors generated from the vectorial labels.
The resulting complexity of Algorithm~\ref{alg:FeatureMap} is $O(n(m \log \rho + hHD)$, 
while the complexity of Algorithm~\ref{alg:ODDCLkernel} drops to $O(nhHD)$.
Note that, in practice, the computational gain due to approximation might be higher than what the worst-case analysis suggests (see Section~\ref{sec:experiments}), because the worst-case scenarios of the two versions of the algorithm are different. 
 In the approximated case, the complexity is independent from the number of continuous attributes associated with a discrete feature, and the worst case is the one that maximizes the number of discrete features.

\section{Experiments}\label{sec:experiments}
\begin{table*}[h]
\centering
\caption{Average accuracy results $\pm$ standard deviation in nested 10-fold cross validation of the proposed ODDCL$_{ST}$, GraphHopper, Shortest Path, Common Subgraph Matching, Propagation, Graph Invariant, Weisfeiler-Lehman and ODD$_{ST}$ kernels. The last two kernels do not consider information from node attributes. The best accuracy for each dataset is reported in bold.}\label{tab:nestedkfoldresults}
   \begin{tabular}{|l|c|c|c|c|c|c|}
\hline
$Kernel$ 		&COX2 & BZR & DHFR &  ENZYMES 		& PROTEINS 		& SYNTH \\
\hline
ODDCL$_{ST}$ &\textbf{83.01$\pm$1.95} & \textbf{92.85$\pm$0.30}& \textbf{84.05$\pm$3.02}& 73.40$\pm$0.98 &75.86 $\pm$ 0.77  & 95.83$\pm$ 0.73 \\
 ODDCL$_{Approx}$ & 79.99$\pm$1.20&87.12$\pm$1.14 & 81.48$\pm$0.36& 69.95 $\pm$ 1.12 & \textbf{75.93 $\pm$ 0.59} & 96.1 $\pm$ 0.74 \\
GH	& -& -&- & 68.3$\pm$1.31 &	74.1$\pm$0.5 &78.3$\pm$1.5 \\
SP		& -& -& -& 72.3$\pm$0.9 & 75.5$\pm$0.8 &82.5$\pm$1.3\\
CSM		& -& -&- & 69.5$\pm$0.7& out of memory&out of time\\
P2K & 77.41$\pm$0.65 & 82.06$\pm$0.51 & 67.23$\pm$0.84& 67.99&69.37$\pm$0.81 & 75.16$\pm$1.37\\
\hline
GIK\_NSPDK &71.68$\pm$0.99	&86.84$\pm$0.55&	70.76$\pm$1.07 & 76.18$\pm$0.52 & 70.78$\pm$0.91&83.46$\pm$1.38  \\
GIK\_SGK\_GLOBAL & 78.09$\pm$0.58&	70.80$\pm$0.96	&81.41$\pm$0.65 & 76.53$\pm$0.53	&	71.38$\pm$0.53	&83.43$\pm$1.41\\
GIK\_SGK\_LOCAL & 81.07$\pm$0.82& 86.23$\pm$0.67&	81.24$\pm$0.66 & 76.22$\pm$0.40&	71.40$\pm$0.60&83.83$\pm$1.31\\
GIK\_WL\_GLOBAL & 81.63$\pm$0.98&	69.11$\pm$0.93& 81.46$\pm$0.84 & 76.17$\pm$0.48&70.21$\pm$0.88&82.66$\pm$1.02 \\
GIK\_WL\_LOCAL & 71.66$\pm$0.97&	86.54$\pm$0.85& 71.66$\pm$0.97 & \textbf{76.62$\pm$0.63}	&	71.85$\pm$0.71&82.60$\pm$0.77\\
\hline
WL		& 81.45$\pm$0.76&88.41$\pm$0.42 & 82.76$\pm$0.41& 48.5$\pm$0.7 & 75.6$\pm$0.5 &\textbf{97.5$\pm$2.4}\\
ODD$_{ST}$ &82.76$\pm$0.97 & 88.21 $\pm$ 0.54& 83.75 $\pm$ 1.03&  51.26$\pm 1.03$&73.85$\pm 0.55$ &96.63$\pm 0.64$ \\
\hline
\end{tabular}
 \end{table*}
In this section, we compare the ODDCL$_{ST}$ kernel presented in Section~\ref{sec:stcontinuous} 
and its approximated version presented in Section~\ref{sec:approximation}
with several state-of-the-art kernels for graphs with continuous labels. 
After the description of the experimental setup in Section \ref{sec:exsetup}, we discuss the predictive performance of the different kernels in Section~\ref{sec:experimentalresults}. In Section~\ref{sec:computationaltimes} we compare the computational times required by the different kernel calculations.
\subsection{Experimental setup}\label{sec:exsetup}
The experimental setup follows the one in~\cite{Feragen2013}: the results presented in this section refer to an SVM classifier\footnote{https://github.com/nickgentoo/scikit-learn-graph} in a process of nested 10-fold cross validation, in which the kernel (and the classifier) parameters are validated using the training dataset only; the experiments have been repeated 10 times (with different cross validation splits), and the average accuracy results with standard deviation are reported. 
The proposed kernel parameter values have been cross validated from the following sets: $h$=$\{0,1,2,3\}$, $\lambda$=$\{0.1, 0.3, 0.5, 0.7, 1, 1.2\}$.
For ODDCL$_{Approx}$ the $D$ parameter has been selected after a preliminary experiment: a nested 10-fold cross validation on the Enzymes dataset was performed using the default values for all other parameters and varying only $D$; $D=1000$ looked like a good trade off between speed and accuracy. 
Note that, with very high values of $D$, it is expected to reproduce almost exactly the results of the ODDCL$_{ST}$ kernel. 
The SVM $C$ parameter has been selected in the set $\{0.01, 0.1, 1, 10, 100, 1000, 10000\}$.
All the considered kernels depend on node kernels for continuous vectors and/or discrete node labels. The kernel for continuous attributes has been fixed for all the kernels as $K_A(v_1,v_2)= e^{-\lambda ||A(v_1)-A(v_2)||^2}$ with $\lambda=1/d$ and $d$ the size of the vector of attributes $A(v_i)$.
Where discrete labels were not available, the degree of each node has been considered as the node label.
{Note that this step is not strictly necessary, since if the graphs have no discrete label information, we can just assume all the nodes having the same label. However, agreement on out-degree is an effective way to speed-up computation and increase the discriminativeness of kernels.}
Following~\cite{Feragen2013}, the kernel matrices have been normalized.

We tested our method on the (publicly available) datasets from~\cite{Feragen2013}: \textit{ENZYMES}, \textit{PROTEINS} and \textit{SYNTHETIC}, and from \cite{Neumann2015}: COX2, BZR, DHFR. 
\textit{ENZYMES} (symmetrized version) is a set of proteins from the BRENDA database~\cite{Schomburg2004}.
Each protein is represented as a graph, where nodes correspond to secondary structure elements. Two nodes are connected whenever they are neighbors either in the amino acid sequence or in the 3D space of the protein tertiary structure~\cite{Borgwardt2005a}.
Each node has a discrete attribute indicating the structure it belongs to (helix, sheet or turn).
Moreover, several chemical and physical measurements can be associated with each node, obtaining a vector-valued attribute associated to each node. Examples of these measurements are the length of the secondary structure element in \AA, its hydrophobicity, polarity, polarizability etc.
The task is the classification of enzymes into one out of 6 EC top-level classes. There are 100 graphs per class in the dataset. The average number of nodes and edges of the graphs is $32.6$ and $46.7$, respectively. The size of the vectors associated with the nodes is 18.\\
\textit{PROTEINS} is the dataset from \cite{dobson2003}. The proteins are represented as graphs as described above. The task is to distinguish between enzymes and non-enzymes. There are 1113 graphs in the dataset, each one with an average of $39.1$ nodes and $72.8$ edges. The dimension of the continuous node attributes is 1.\\
\textit{SYNTHETIC} is a dataset presented in~\cite{Feragen2013}. A random graph with 100 nodes and 196 edges has been generated. Each node has a corresponding continuous label sampled from $\mathcal{N}(0,1)$. Then 2 classes of graphs have been generated with 150 graphs each. Each graph in the first class was generated rewiring 5 edges and permuting 10 node attributes from the original graph, while for each graph in the second class 10 edges and 5 node attributes has been modified. Finally, noise from $\mathcal{N}(0,0.45^2)$ has been added to each node attribute in both classes.
\textit{COX2, BZR} and \textit{DHFR}, orginally presented in \cite{Vert2009} are datasets of chemical compounds where the target is to predict their toxicity. The 3D coordinates of the atoms have been used as node attributes. COX2 counts 467 graphs, while BZR and DHFR 405 and 756,  respectively. The average number of nodes for the three datasets is $41.2$, $35.7$ and $42.4$, while the average number of edges is $43.44$, $38.35$ and $44.5$, respectively. 
\subsection{Experimental results}\label{sec:experimentalresults}
In Table~\ref{tab:nestedkfoldresults} we report the experimental results of the proposed ODDCL$_{ST}$ kernel, its approximated version ODDCL$_{Approx}$ presented in Section~\ref{sec:approximation}, the Propagation kernel ($P2K$)\cite{Neumann2015}, different instantiations of the Graph Invariant Kernels~\cite{Orsini} (GIK\_NSPDK, GIK\_SGK\_GLOBAL, GIK\_SGK\_LOCAL, GIK\_WL\_GLOBAL, GIK\_WL\_LOCAL)
and the results from the paper \cite{Feragen2013}, corrected according to the \textit{erratum}. 
The results reported from~\cite{Feragen2013} are those of the GraphHopper kernel (GH)~\cite{Feragen2013}, the connected subgraph matching kernel (CSM)~\cite{Kriege2012} and the shortest path kernel (SP)~\cite{Kriegel05shortestpath}. 
All these kernels can deal with continuous labels. For sake of comparison, the results of the Weisfeiler-Lehman kernel (WL)~\cite{NIPS2009_0533} and  ODD$_{ST}$~\cite{Dasan2012} kernels, that can deal only with discrete attributes, are reported too.

In the \textit{ENZYMES} dataset, 
GIK kernels are the best performing ones. The proposed ODDCL$_{ST}$ kernel performs better than GH, SP, CSM and P2K, while its approximated version is below SP in terms of accuracy but still performs better than P2K, GH and CSM. Note however that the computational requirements of SP are prohibitive on this dataset, as it will be detailed later in this section. In this dataset, the WL and the ODD$_{ST}$ kernels perform poorly, indicating that the information encoded by continuous attributes is relevant to the task.
{In \textit{PROTEINS}, WL performs better than the other competitors, suggesting that in this case the continuous labels may be not much informative. 
However, ODDCL$_{ST}$ and ODDCL$_{Approx}$ kernels achieve a slightly higher accuracy than WL. 
This gives further evidence that our proposed kernels are able to extract useful information from non-discrete labels in a more effective way than the other kernels for continuous labels. }
%
Before analyzing the results for the \textit{SYNTHETIC} dataset, we need to draw some considerations.
The first reported results on this dataset were affected by a bug. The correct results have been published later, and depicted a scenario where the kernels that did not consider continuous labels performed better than the others. 
However, it is interesting to evaluate the performance of the different kernels on this dataset because it shows how much a kernel is resistant to noise.
In the \textit{SYNTHETIC} dataset, even if the proposed kernels perform better than other competitors that consider continuous attributes, its performance are not able to achieve the WL and ODD$_{ST}$ ones. This means that, for this dataset, the information provided by vectorial labels is basically noise. This fact is more evident when looking at the performance of the other kernels that consider vectorial labels.
However, it is clear that the proposed ODDCL$_{ST}$ kernel and its approximated version are more tolerant to this kind of noise than competing kernels.
Moreover, for the proposed kernel it is possible to consider an additional parameter, to be cross-validated with the others, that indicates if the vectorial labels have to be considered or not. We recall that, without considering vectorial labels, ODDCL$_{ST}$ kernel reduces to the ODD$_{ST}$ kernel.

On COX2 and DHFR datasets, ODDCL$_{ST}$ is the best performing kernel, while its approximated version is competitive with the other kernels. 
On BZR dataset, ODDCL$_{ST}$ and ODDCL$_{Approx}$ are the first and the second best performing kernels among the ones considering vectorial attributes. 

Summarizing the results, ODDCL$_{ST}$ kernel performs better than the other kernels that consider vectorial labels on five out of six datasets. At the same time, ODDCL$_{Approx}$  is competitive with the other kernels; in particular, it performs always better than GH (where applicable) and P2K. 

{In order to assess the differences in performance among the kernels, we performed 
the Friedman test with the Nemenyi post-hoc test 
($\alpha=0.05$)~\cite{JapkowiczNathalieandShah2011}. Since not all results are available for all kernels on all datasets, we performed one test on all datasets but excluding the kernels GH, SP, CSM; a second test includes all kernels but it is performed on ENZYMES, PROTEINS, SYNTHETIC datasets: ODDCL$_{ST}$ is significantly better than all other kernels except ODDCL$_{Approx}$, ODD$_{ST}$ and WL. Since WL does not take into account continuous labels, we performed a further analysis focusing on the datasets for which such information is important: COX2, BZR, DHFR, ENZYMES. On such datasets ODDCL$_{ST}$ always outperforms WL; we performed a t-test~\cite{JapkowiczNathalieandShah2011} which showed that ODDCL$_{ST}$ is significantly better than WL at level $\alpha=0.05$ on BZR and ENZYMES. }
\subsection{Computational times}\label{sec:computationaltimes}
\begin{table}[t]
\caption{Time required for the kernel matrix computation of ODDCL$_{ST}$, ODDCL$_{Approx}$, GraphHopper, Shortest Path, Propagation and Graph Invariant kernels.$*$: the times referring to GH and SP are reported from \protect\cite{Feragen2013}. Notation d: day(s); h: hour(s), m: minute(s); s:second(s) }\label{tab:times}
\centering
\begin{tabular}{|l|c|c|c|}
\hline
$Kernel$ 		&  ENZYMES 		& PROTEINS 		& SYNTH \\
\hline
ODDCL$_{ST}$	& 35m& 21m& 23m\\
ODDCL$_{Approx}$ & 1.8m & 5.2m & 7.5m \\ 
GH$^*$ &12m10s&2.8h&12m10s\\
SP$^*$ &3d&7.7d&3.4d\\
GIKs (average) & 48m & 3h49m & 16.3m \\
P2K &6s &28s &5s \\
\hline
\end{tabular}
\end{table}
The computational times reported in this section refer to the Gram matrix computation of each kernel on the three largest datasets. 
Since the kernels are implemented in different languages, we considered for sake of comparison the computational times reported in~\cite{Feragen2013} for GH and SP kernels.
We want to point out that the times reported in this section has to be considered just as orders of magnitude.
Table~\ref{tab:times} reports such computational times for the different kernels.
P2K is the fastest kernel, however its predictive performances are the poorest. 
{
The proposed ODDCL$_{ST}$ kernel is faster than the SP kernel in all the considered datasets. With respect to GraphHopper, ODDCL$_{ST}$ is slightly slower on ENZYMES and SYNTH (roughly two times slower, but still in the order of minutes), while it is considerably faster on the PROTEINS dataset (eight times faster).
}
{The computational requirements of ODDCL$_{ST}$ and GIKs are comparable on ENZYMES and SYNTH datasets, while ODDCL$_{ST}$ is considerably faster on the PROTEINS dataset (almost eleven times faster)}.
{Let us now consider} the approximated version {of our proposed kernel}, ODDCL$_{Approx}$. It is the second fastest kernel (after P2K), with a significant difference with respect to GraphHopper while being more accurate in all the considered datasets. ODDCL$_{Approx}$ is also considrably faster than GIKs, being more accurate in two out of three of the considered datasets. 
Interestingly, because of the different substructures considered by the two kernels, GraphHopper kernel have the lowest run-times in ENZYMES and SYNTHETIC, while for ODDCL$_{ST}$ PROTEINS is the dataset that requires the lowest computational resources.
We can argue that this happens because of the higher number of edges of proteins, that directly influence the sparsity of ODDCL$_{ST}$ kernel.
On the other hand, ODDCL$_{Approx}$ shows the maximum speedup on the ENZYMES dataset. The speedup obtained by the approximated version is proportional to the number of different discrete features generated by the ODD base kernel, and thus is influenced by the number of different discrete labels in the dataset, in addition to the graph topology.

\section{Conclusions}\label{sec:conclusions}
In this paper, we have presented an extension to continuous attributes of the  ODD kernel framework for graphs.
Moreover, we have studied the performances of a continuous attributes graph kernel derived by the ST kernel for trees.
Experimental results on reference datasets show that the resulting kernel is both fast to compute and quite effective
on all studied datasets, which is not the case for  continuous attributes graph kernels presented in literature. A faster but approximated version of the proposed kernel, returning satisfying predictive performances, is presented as well.
\section*{Acknowledgements}
This work was supported by the University of Padova under the strategic project BIOINFOGEN.
\bibliographystyle{IEEEtran}
\bibliography{mendeley}

\begin{thebibliography}{10}
\providecommand{\url}[1]{#1}
\csname url@samestyle\endcsname
\providecommand{\newblock}{\relax}
\providecommand{\bibinfo}[2]{#2}
\providecommand{\BIBentrySTDinterwordspacing}{\spaceskip=0pt\relax}
\providecommand{\BIBentryALTinterwordstretchfactor}{4}
\providecommand{\BIBentryALTinterwordspacing}{\spaceskip=\fontdimen2\font plus
\BIBentryALTinterwordstretchfactor\fontdimen3\font minus
  \fontdimen4\font\relax}
\providecommand{\BIBforeignlanguage}[2]{{%
\expandafter\ifx\csname l@#1\endcsname\relax
\typeout{** WARNING: IEEEtran.bst: No hyphenation pattern has been}%
\typeout{** loaded for the language `#1'. Using the pattern for}%
\typeout{** the default language instead.}%
\else
\language=\csname l@#1\endcsname
\fi
#2}}
\providecommand{\BIBdecl}{\relax}
\BIBdecl

\bibitem{Borgwardt2005a}
K.~M. Borgwardt, C.~S. Ong, S.~Sch\"{o}nauer, S.~V.~N. Vishwanathan, A.~J.
  Smola, and H.-P. Kriegel, ``{Protein function prediction via graph
  kernels.}'' \emph{Bioinformatics (Oxford, England)}, vol. 21 Suppl 1, pp.
  47--56, Jun. 2005.

\bibitem{Karaman2014}
S.~Karaman, L.~Seidenari, and S.~Ma, ``{Adaptive Structured Pooling for Action
  Recognition},'' \emph{BMVA}, pp. 1--12, 2014.

\bibitem{Kriegel05shortestpath}
K.~Borgwardt and H.-P. Kriegel, ``{Shortest-Path Kernels on Graphs},'' in
  \emph{ICDM}, vol.~0.\hskip 1em plus 0.5em minus 0.4em\relax Los Alamitos, CA,
  USA: IEEE, 2005, pp. 74--81.

\bibitem{Dasan2012}
G.~{Da San Martino}, N.~Navarin, and A.~Sperduti, ``{A Tree-Based Kernel for
  Graphs},'' in \emph{Proceedings of the Twelfth SIAM International Conference
  on Data Mining}, 2012, pp. 975--986.

\bibitem{Gartner2003a}
T.~Gartner, P.~Flach, S.~Wrobel, and T.~G\"{a}rtner, ``{On Graph Kernels:
  Hardness Results and Efficient Alternatives},'' in \emph{16th Annual
  Conference on Computational Learning Theory and 7th Kernel Workshop}, ser.
  LNCS, vol. 2777.\hskip 1em plus 0.5em minus 0.4em\relax Springer Berlin
  Heidelberg, 2003, pp. 129--143.

\bibitem{Costa2010}
F.~Costa and K.~{De Grave}, ``{Fast neighborhood subgraph pairwise distance
  kernel},'' in \emph{ICML}.\hskip 1em plus 0.5em minus 0.4em\relax Omnipress,
  2010, pp. 255--262.

\bibitem{NIPS2009_0533}
N.~Shervashidze and K.~Borgwardt, ``{Fast subtree kernels on graphs},'' in
  \emph{NIPS}, 2009, pp. 1660--1668.

\bibitem{Birlinghoven}
M.~Neumann, N.~Patricia, R.~Garnett, and K.~Kersting, ``{Efficient Graph
  Kernels by Randomization},'' in \emph{ECML PKDD}, ser. LNCS, vol. 7523.\hskip
  1em plus 0.5em minus 0.4em\relax Springer Berlin Heidelberg, 2012, pp.
  378--393.

\bibitem{Kriege2012}
N.~Kriege and P.~Mutzel, ``{Subgraph matching kernels for attributed graphs},''
  in \emph{ICML}, 2012, pp. 1015--1022.

\bibitem{Feragen2013}
A.~Feragen, N.~Kasenburg, J.~Petersen, M.~de~Bruijne, and K.~M. Borgwardt,
  ``{Scalable kernels for graphs with continuous attributes},'' in \emph{NIPS},
  2013, pp. 216--224.

\bibitem{Neumann2015}
M.~Neumann, R.~Garnett, C.~Bauckhage, and K.~Kersting, ``{Propagation kernels:
  efficient graph kernels from propagated information},'' \emph{Machine
  Learning}, Jul. 2015.

\bibitem{Orsini}
F.~Orsini, P.~Frasconi, and L.~D. Raedt, ``{Graph invariant kernels},'' in
  \emph{IJCAI}, 2015.

\bibitem{DBLP:journals/ijon/MartinoNS16}
G.~D.~S. Martino, N.~Navarin, and A.~Sperduti, ``{Ordered Decompositional DAG
  kernels enhancements},'' \emph{Neurocomputing}, vol. 192, pp. 92--103, 2016.

\bibitem{Moschitti2006}
A.~Moschitti, ``{Making Tree Kernels Practical for Natural Language
  Learning},'' in \emph{Proceedings of the 11th Conference of the European
  Chapter of the Association for Computational Linguistics}, 2006.

\bibitem{DBLP:conf/nips/ViswanathanS02}
S.~V.~N. Vishwanathan and A.~J. Smola, ``{Fast Kernels for String and Tree
  Matching},'' in \emph{NIPS}, 2002, pp. 569--576.

\bibitem{NIPS2008_3495}
A.~Rahimi and B.~Recht, ``{Weighted Sums of Random Kitchen Sinks: Replacing
  minimization with randomization in learning},'' in \emph{NIPS}, 2009, pp.
  1313--1320.

\bibitem{Williams2001}
\BIBentryALTinterwordspacing
C.~Williams and M.~Seeger, ``{Using the Nystr{\"{o}}m Method to Speed Up Kernel
  Machines},'' in \emph{Advances in Neural Information Processing Systems 13},
  2001, pp. 682--688. [Online]. Available:
  \url{http://citeseerx.ist.psu.edu/viewdoc/summary?doi=10.1.1.18.7519}
\BIBentrySTDinterwordspacing

\bibitem{Schomburg2004}
I.~Schomburg, A.~Chang, C.~Ebeling, M.~Gremse, C.~Heldt, G.~Huhn, and
  D.~Schomburg, ``{BRENDA, the enzyme database: updates and major new
  developments.}'' \emph{Nucleic Acids Res.}, vol.~32, pp. D431--D433, 2004.

\bibitem{dobson2003}
P.~D. Dobson and A.~J. Doig, ``{Distinguishing Enzyme Structures from
  Non-enzymes Without Alignments},'' \emph{Journal of Molecular Biology}, vol.
  330, no.~4, pp. 771--783, 2003.

\bibitem{Vert2009}
P.~M. J.-p. Vert, ``{Graph kernels based on tree patterns for molecules},''
  \emph{Machine Learning}, no. September 2008, pp. 3--35, 2009.

\bibitem{JapkowiczNathalieandShah2011}
N.~Japkowicz and M.~Shah, \emph{Evaluating Learning Algorithms: A
  Classification Perspective}.\hskip 1em plus 0.5em minus 0.4em\relax New York,
  NY, USA: Cambridge University Press, 2011.

\end{thebibliography}

\end{document}